# SceneGTMM: A Conformal Mapping-based Scene-Aware Transferable GNN-Transformer Dual-Graph Interaction Framework for Map Matching

Yongliang Zhang[†‡]
zhangyongliang.zha@alibaba-inc.com
Amap, Alibaba Group,beijing,china

Feng Song
feng.song@alibaba-inc.com
Amap, Alibaba Group,beijing,china

Ji Chen
cj270839@alibaba-inc.com
Amap, Alibaba Group,beijing,china

Lishuai Guo[†]
lishuai.gls@alibaba-inc.com
Amap, Alibaba Group,beijing,china

Yong Deng
dy193534@autonavi.com
Amap, Alibaba Group,beijing,china

Yue Zheng
heinstein.zy@autonavi.com
Amap, Alibaba Group,beijing,china

Tianyi Liu
liutianyi.lty @alibaba-inc.com
Amap, Alibaba Group,beijing,china

Zhixiong Chen
jeremy.czx@autonavi.com
Amap, Alibaba Group,beijing,china

Qixin Zhang
qixin.zqx@alibaba-inc.com
Amap, Alibaba Group,beijing,china

## Abstract

Map matching is a pivotal technology bridging positioning data and high-precision road networks, with core challenges residing in noise robustness within dynamic environments, cross-regional transferability, and model interpretability. To address the limitations of existing methods regarding local-global information fusion, dynamic road network adaptation, and reliance on black-box models, this paper proposes SceneGTM, a transferable GNN-Transformer dual-graph interaction map matching framework based on a Conformal Mapping-based Scene-Relative Strategy. The framework achieves breakthroughs through three key innovations: 1) Conformal Mapping Scene-Relative Strategy: Dynamically constructs trajectory-centric local coordinate systems to decouple the model from bindings to the training road network, enabling cross-regional transferability and dynamic road network updates; 2) GNN-Transformer Dual-Graph Interaction Architecture: A Road Graph (modeled by GNN) captures local topological constraints, while a Trajectory Graph (modeled by Transformer) captures global temporal dependencies, achieving noise suppression and semantic alignment via a cross-graph attention mechanism; 3) CRF-Enhanced Structured Prediction: Combines the global context awareness of the Transformer with the topological transition constraints of Conditional Random Fields (CRF) to enhance path connectivity and prediction robustness. Experiments demonstrate that SceneGTM maintains an accuracy of over 80% across multi-source trajectory data (with positioning errors ranging from 16 to 50 meters), representing a 5.3% improvement over traditional HMM. Furthermore, it outperforms existing deep learning methods (e.g., MTrajRec, GraphMM, TMM-LGD) in cross-city transfer scenarios, validating its dynamic adaptability and generalization capabilities. Additionally, the model significantly enhances decision interpretability through the visualization of attention weights and relative coordinate transformations. This study provides a new paradigm for high-precision, transferable map matching for real-time traffic perception and autonomous driving path planning.

## CCS Concepts

• Information systems → Geographic information systems; • Information systems → Location based services; • Computing methodologies → Neural networks; • Computing methodologies → Supervised learning by classification; • Theory of computation → Sequential decision making; • Mathematics of computing → Combinatorial optimization.

## Keywords

Map Matching,Conformal Mapping,GNN-Transformer,
Dual-Graph Interaction,Transferability,Noise Robustness

## 1 Introduction

Map matching is an algorithmic process that correlates raw positioning data (e.g., coordinates, speed, heading) collected by mobile devices (such as smartphones and onboard terminals) with road networks in high-precision digital maps to determine the most likely path traveled by the device within the road network. Its core objective is to correct positioning errors (such as trajectory drift), "snap" discrete coordinate points onto actual roads, and infer accurate travel trajectories, as illustrated in Figure 1. Map matching serves as a bridge connecting the physical space with the digital world; its value lies in transforming "noisy data" into an actionable basis for intelligent decision-making. With the advancement of 5G, V2X (Vehicle-Infrastructure Cooperation), and AI technologies, map matching is evolving from "post-processing" to "real-time interaction," becoming critical infrastructure for smart cities and the autonomous driving ecosystem.

The following section provides a classified summary based on the development of map matching algorithms.

(1) Traditional Methods

Based on geometry and topology, relying on hand-crafted rules and prior knowledge.

- Geometric Methods: Select the nearest road by calculating the spatial distance between trajectory points and road segments (e.g., Euclidean distance, perpendicular distance), such as nearest neighbor search and projection matching. This method is sensitive to noise; trajectory point errors (e.g., due to multipath effects) can easily lead to matching incorrect roads, especially in dense road networks. It ignores topological constraints and road connectivity, potentially generating unreasonable paths (e.g., road-crossing jumps).
- Topological Methods: Incorporate road network connectivity, optimizing path selection through shortest path algorithms or dynamic programming, such as shortest path matching based on Dijkstra's algorithm. However, this method overly relies on topological structures, assuming drivers choose the shortest path, whereas in reality, they may select non-optimal routes due to preferences (e.g., avoiding congestion). It exhibits poor dynamic adaptability and cannot handle real-time traffic variations.

(2) Probabilistic Model Methods

For example, Hidden Markov Models (HMM) [1,2,3], which infer paths by maximizing the joint emission and transition probabilities. This method assumes trajectory point errors follow a fixed distribution (e.g., Gaussian), whereas in reality, noise is spatially heterogeneous due to urban density and building layouts. Emission and transition probabilities require manual design, making it difficult to capture complex patterns (e.g., driver behavior preferences). HMM relies on sequential decision-making, which may lead to locally optimal paths.

(3) Sequence-to-Sequence Models

Literature [4]、[5]、[6] encodes trajectory sequences into hidden states and decodes them into road segment sequences. Through a multi-task sequence-to-sequence learning architecture, it synchronously predicts road segments and movement ratios, proposing constraint masks, attention mechanisms, and attribute modules to overcome the limitations of coarse-grained grid representations and improve performance. However, this method suffers from long-sequence dependency issues, making it difficult to capture global context in long-distance trajectories (e.g., cross-regional path planning). It is sensitive to sparse data; low-sampling-rate trajectories (sparse points) can easily lead to decoding errors. Furthermore, the model training is bound to the road network of the training set, preventing dynamic updates and transferable inference.

(4) Sequence-to-Sequence Models + Graph Neural Networks

Literature [7] leverages the graph characteristics of map matching, combining Graph Neural Networks (GNN) with conditional models. It fully utilizes the graph topological structures of road networks and trajectories, aligning road segments and trajectories in a latent space. This method faces a sparse trajectory bottleneck; low-sampling-rate trajectories lead to missing graph structural information, affecting matching accuracy. Furthermore, since the trajectory-augmented graph is bound to the training set road network, it cannot achieve dynamic updates or transferable inference.

(5) Transformer Architecture

Literature[9]、[10]utilizes the Self-Attention mechanism to model global dependencies between trajectory points and roads. It designs a Transformer-based encoder-decoder map matching framework that automatically learns context representations of noisy trajectory points in an end-to-end manner, inferring driver trajectory behavior and road network structures. This method has several advantages: global context modeling allows the attention mechanism to capture long-distance dependencies, adapting to complex road networks; end-to-end optimization automatically learns noise patterns and driver preferences without requiring hand-crafted features. However, the disadvantages are also evident: poor interpretability makes it difficult to analyze decision logic in black-box models, complicating debugging and optimization. The model training is bound to the training set road network, preventing generalization to other regions, and similarly failing to achieve dynamic updates or transferable inference.

To address the aforementioned issues, the following core contradictions must be resolved:

- Balance between Local and Global Information: Traditional methods focus only on local geometric/topological constraints, while deep learning methods (e.g., Transformer) emphasize global dependencies but lack fine-grained modeling of local road features..
- Dynamic Updates and Cross-Regional Transfer: Existing models are bound to the training road network, making it difficult to adapt to dynamic road networks or cross-regional scenarios.
- Noise Robustness and Interpretability: The black-box nature of models like Transformer hinders the debugging and optimization of decision logic.

Based on the above analysis, this paper proposes SceneGTMM: A Transferable GNN-Transformer Dual-Graph Interaction Map Matching Framework based on a Scene-Relative Strategy. SceneGTMM employs a scene-relative strategy and dual-graph interaction architecture, featuring transferable and lightweight designs with enhanced interpretability. It addresses the bottlenecks of existing methods in transferability, dynamic adaptability, and interpretability, providing a new paradigm for real-time, high-precision map matching.

(1) Scene-Relative Strategy: Solving Transferability and Dynamic Updates

- Core Idea: Decouple the model from the training road network by dynamically constructing scene-relative features.
- Trajectory-Road Relative Graph: During inference, subgraphs are dynamically constructed based on the local road structure of the current trajectory, rather than relying on the global road network.
- Conformal Mapping: Maps the coordinates of trajectory points and road segments to a trajectory-centric local coordina-

te system, reducing the impact of absolute position deviations.

(2) GNN-Transformer Dual-Graph Interaction: Fusing Local and Global Information

- Dual-Graph Modeling: The Road Graph (GNN) uses Graph Neural Networks to model the topological structure of the road network, capturing local road features (e.g., connectivity, speed limits). The Trajectory Graph (Transformer) models global temporal dependencies of trajectory points through self-attention mechanisms, adapting to long-distance path planning. The GNN provides precise local road constraints, while the Transformer models global path preferences; cross-graph attention suppresses noise interference caused by multipath effects.
- Cross-Graph Attention Mechanism: Road graph node embeddings are introduced into the Transformer decoder to dynamically align the semantic space of trajectory points and road segments. By visualizing cross-graph attention weights, the association strength between trajectory points and road segments can be analyzed.

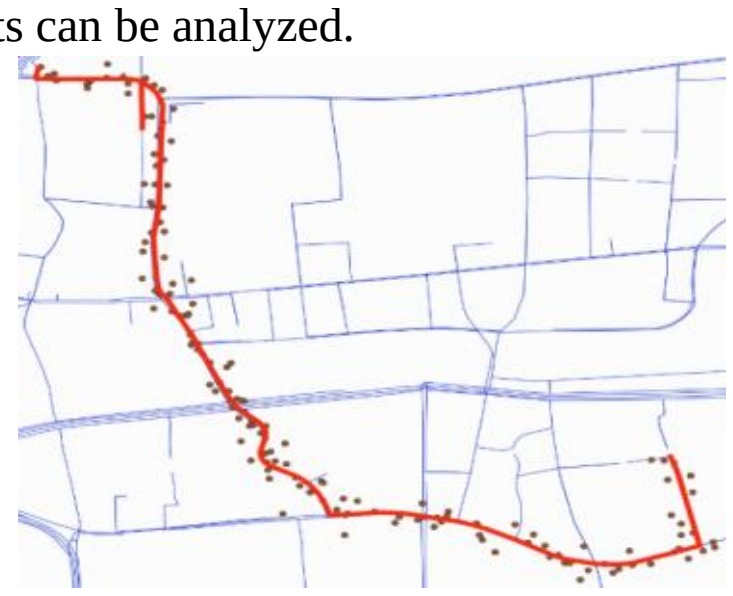

Figure 1: Map Matching

# 2 Methodology

## 2.1 Road Network Conformal Mapping for Scene-Relative Processing

Road Network Scene-Relative Processing is a key preprocessing paradigm for achieving high-precision, transferable map matching. Its core value lies in employing relative coordinate transformation, dynamic spatial normalization, and local topological modeling to transform the map matching problem from a traditional "ID-dependent" paradigm to a "scene-aware" one. This thereby provides structured input for the GNN-Transformer hybrid architecture, ultimately achieving cross-city and cross-road network generalization capabilities (e.g., a model trained in Beijing can be directly applied to the Guangzhou road network).

*2.1.1* ***Relevant Road Network Extraction.*** In this paper, the road network relevant to the trajectory points to be matched is extracted based on the buffer range r of the trajectory points (recommended within 50 meters), as shown in Figure 2, and continuous IDs are assigned in order to achieve road network pruning preprocessing. Road network pruning is an indispensable preprocessing paradigm in map matching. Road network pruning based on trajectory point buffers is a dynamic spatial filtering strategy, whose core function is to improve matching efficiency and accuracy by localizing the road network search space, while enhancing the model's adaptability to complex road network scenarios. Its essence lies in improving the model's learning ability for key structures while reducing computational costs through topological simplification and feature focusing. In transferable map matching (e.g., cross-city applications), pruning provides efficient and robust input representations for the GNN-Transformer hybrid architecture by eliminating redundant dependencies and reinforcing general topological features.

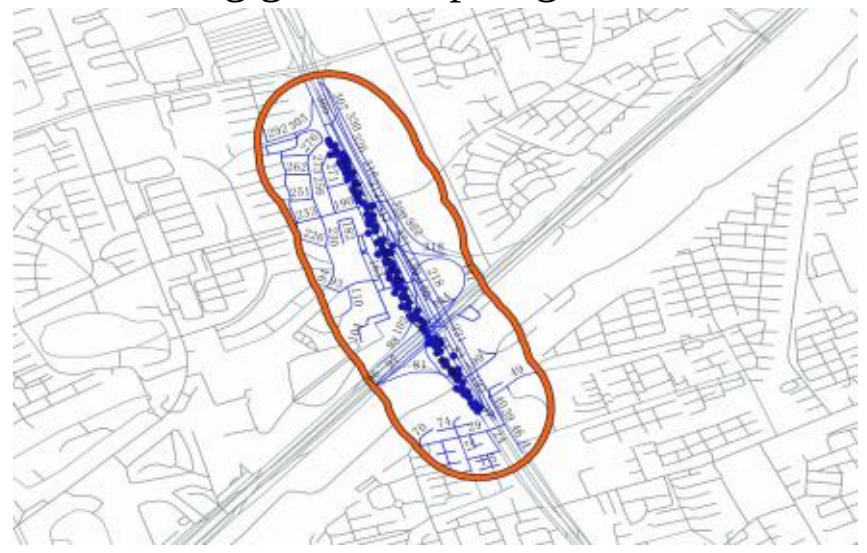

Figure 2: Road Network Scene Extraction

*2.1.2* ***Conformal Mapping Scene-Relative Processing.*** Conformal mapping[11、12] is a transformation that locally preserves angles, serving as a bridge connecting geometry and analysis within complex analysis. Through its elegant "angle-preserving" property, it simplifies many physical and engineering problems. With a profound theoretical foundation (e.g., the Riemann Mapping Theorem) and wide-ranging applications, it is an indispensable tool in modern mathematics and applied sciences. In this paper, complex road networks are mapped onto a unit disk via conformal mapping (e.g., Schwarz–Christoffel, Circle Packing, or Finite Element Conformal Mapping) to achieve scene-relative processing, as shown in Figure 3. Due to the central symmetry of the circle, this approach offers potential advantages both theoretically and practically in trajectory point matching tasks, helping machine learning models capture matching features more efficiently.

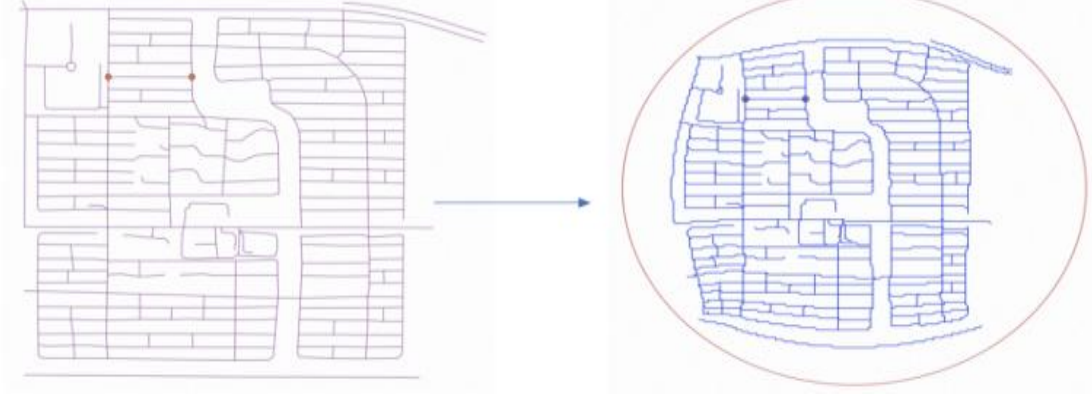

Figure 3: Road Network Conformal Mapping Scene-Relative Processing

### A. *Conformal Mapping from Unit Disk to Rectangular Region*

The Schwarz-Christoffel[13、14] mapping conformally maps the unit disk D to a simply connected polygonal region P with n vertices, whose standard form is:

$$f(w) = A + C \int_0^w \prod_{k=1}^{n} \left(1 - \frac{\zeta}{w_k}\right)^{\alpha_k - 1} d\zeta \quad (1)$$

Where:

- $z_k = f(w_k)$are the vertices of the polygon, and $w_k \in \partial\mathbb{D}$(i.e.,$|w_k| = 1$)are the corresponding pre-vertices;
- The interior angle of the polygon at $z_k$is $\alpha_k\pi$,satisfying$0 < \alpha_k < 2$ (for convex polygons, $0 < \alpha_k < 1$);
- $C \in \mathbb{C} \setminus \{0\}$ controls scaling and rotation;
- $A \in \mathbb{C}$ controls translation;
- The integration path lies within the unit disk and avoids boundary singularities.

***B. Conformal Mapping from Rectangular Region to Unit Disk***

Solving the mapping from interior points of Rectangular Region to the unit disk[15] is essentially finding the inverse of the Schwarz–Christoffel (SC) mapping. This is equivalent to finding the inverse function $\phi = f^{-1}$of the standard SC mapping$f: \mathbb{D} \to P$. Since $\phi$ typically does not have an analytic inverse, it is solved numerically using Newton's downhill method[16].

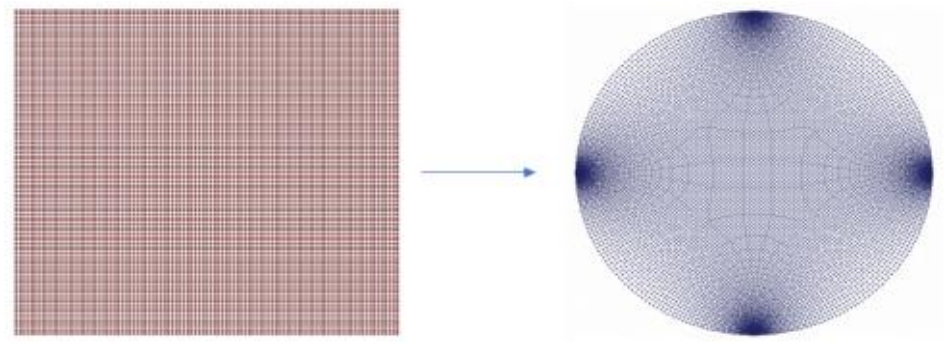

**Figure 4: Mapping from Interior Points of Rectangular Region to Unit Disk**

Using $w_0 = 0$ (the center of the unit disk) for all points is acceptable for mapping the central region, but convergence becomes extremely slow or even fails near the boundaries or corners. Based on the relative position of z within the rectangle, the initial position of w is estimated. Given the topological correspondence between the rectangle and the unit disk, a coarse but effective approximate mapping is established as the initial guess. Core Idea: Normalize the rectangle to [−1,1]×[−1,1][−1,1]×[−1,1], and then map it to the unit disk using bilinear interpolation. The overall algorithm is as follows:

**Algorithm 1:** SC Inverse Mapping via Newton's Method

**Input:** Target point $z$ in the rectangle, SC parameters $\{w_k, \alpha_k\}$, tolerance $\epsilon$.
**Output:** Pre-image $w$ in the unit disk.

```
// (1) Initial Guess w_0
1  Normalize the rectangle to [−1,1] × [−1,1];
2  Estimate initial position w_0 using bilinear interpolation from the
   normalized rectangle to the unit disk;
3  n ← 0;
// (2) Iteration Loop
4  while |f(w_n) − z| ≥ ϵ do
5      Compute f(w_n);
6      Compute derivative f′(w_n) = C ∏_{k=1}^{N} (1 − w_n/w_k)^{α_k − 1};
       // Compute Newton step
7      Δw ← −(f(w_n) − z)/f′(w_n);
       // Backtracking line search
8      Select largest λ ∈ {1, 1/2, 1/4, ...} such that:
9          1. |w_n + λΔw| < 1 (remain within unit disk);
10         2. |f(w_n + λΔw) − z| < |f(w_n) − z| (function value descent);
11     Update w_{n+1} ← w_n + λΔw;
12     n ← n + 1;
13 return w_{n+1};
```

## 2.2 GNN-based Dual-Graph Interactive Embedding of Road Network Scenes and Corresponding Trajectories

We propose a trajectory and road network interactive modeling method that achieves structured fusion of trajectory data and the road network through the construction of a road network-trajectory association matrix and outlier detection. This method offers the following advantages:

- Structured Representation: Trajectories and the road network are mapped into a unified space via road segment buffers, supporting Graph Neural Network (GNN) modeling;
- Dynamic Weight Modeling: The normalized weights of the association matrix establish the dependency of trajectory points on road segments and quantify the dependency strength;
- Robustness Design: The outlier detection mechanism effectively identifies abnormal trajectories, enhancing the model's generalization capability.

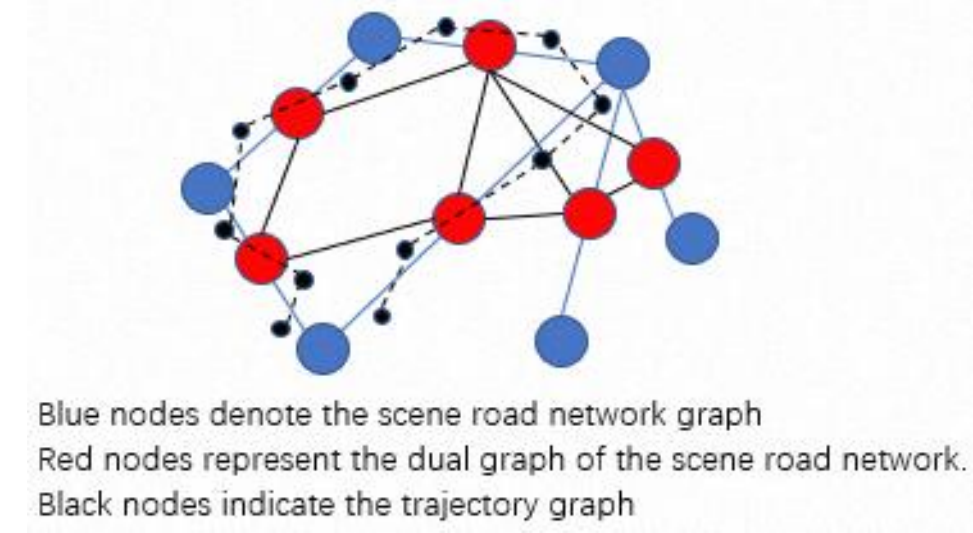

**Figure 5: Dual-Graph Interactive Modeling**

### *2.2.1* ***Dual-Graph Construction and Feature Extraction of Road Network Scenes Following Conformal Mapping Scene-Relative Processing.***

***A. Road Network Scene Dual-Graph Construction***

Traditional road network graphs typically use intersections as nodes and roads as edges; in contrast, the dual graph uses road segments as nodes and the connectivity between adjacent road segments as edges, as shown in Figure 6. This transformation facilitates capturing the direct relationships between road segments, laying the foundation for GNN-based embedding of road network scenes.

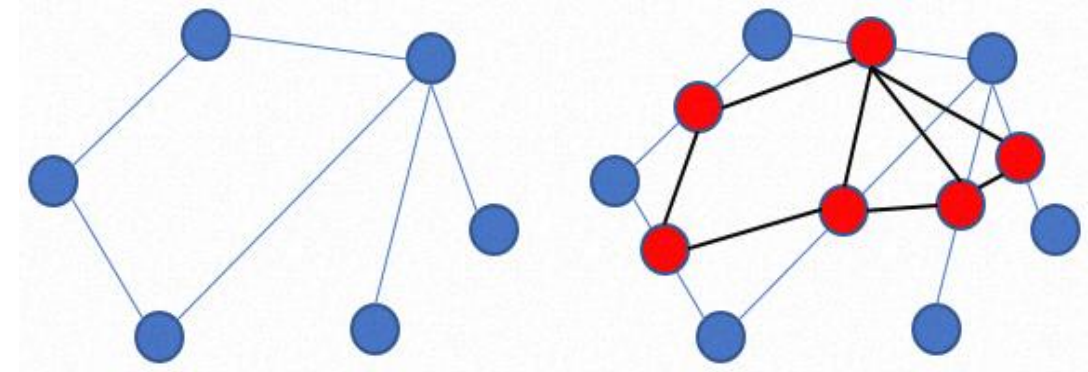

**Figure 6: Road Network Dual-Graph**

***B.Feature Extraction of Road Network Scene Dual-Graph Nodes***

This paper employs the subgraph encoding of VectorNet [17] to extract road segment features of the road network scene (i.e., dual-graph node features). VectorNet, proposed by Waymo et al. in 2020, is a novel method for efficiently modeling high-definition maps and traffic participant dynamics in autonomous driving scenarios. The core idea of this method is to uniformly represent map elements in a vectorized form (i.e., a series of attributed polylines or point sequences). Building on this, it introduces a subgraph and global interaction encoding mechanism to effectively capture local geometric structures and semantic information. Through subgraph encoding, VectorNet achieves efficient, structure-preserving representation learning for vectorized high-definition maps and dynamic agents.

(1) Vectorized Representation of Road Segments

Let road segment r have n nodes, its vectorized representation is shown in Table 1 below. Here, the road segment type is either a road network segment or a trajectory;the node type is represented using one-hot encoding, categorized as start node (1, 0, 0), intermediate node (0, 1, 0), and end node (0, 0, 1).

**Table 1. Vectorized Representation of Road Segments**

| Start Point | | End Point | | Road Type | Node Type | Road ID |
|---|---|---|---|---|---|---|
| $px_0$ | $py_0$ | $px_1$ | $py_1$ | $type_0$ | $role_0$ | $roadId_0$ |
| $px_1$ | $py_1$ | $px_2$ | $py_2$ | $type_1$ | $role_1$ | $roadId_1$ |
| ⋮ | ⋮ | ⋮ | ⋮ | ⋮ | ⋮ | ⋮ |
| $px_{n-1}$ | $py_{n-1}$ | $px_n$ | $py_n$ | $type_{n-1}$ | $role_{n-1}$ | $roadId_{n-1}$ |

(2) VectorNet Subgraph (sub_graph) Encoding of Road Segments

Let road segment r have n nodes. According to the vectorization method in the previous subsection, its vectorized representation is a vectorList containing n−1 9-dimensional vectors. The subgraph encoding process is shown in Figure 7 below. When N=3 (i.e., the sub_graph has three layers), MLP encoding is performed independently on all nodes within the subgraph to extract local features. Through an aggregation operation, all node features are fused to generate the subgraph global representation of the road segment, with a vector dimension of 72.

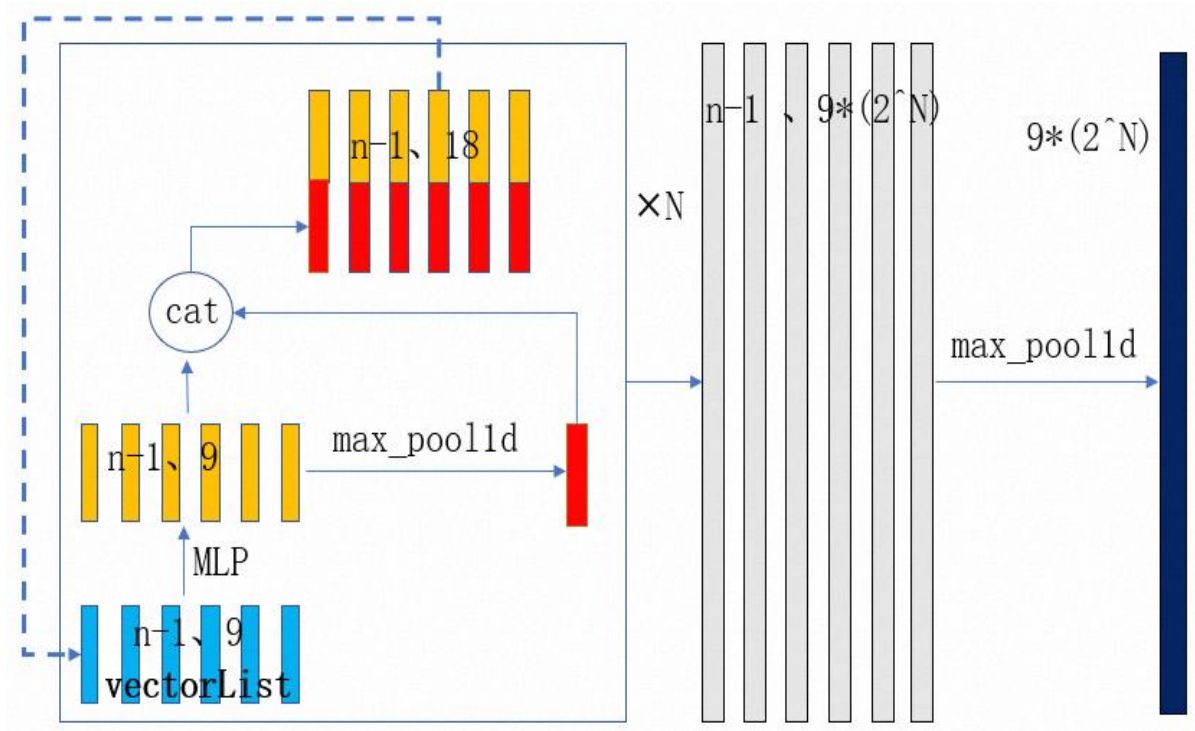

**Figure 7: VectorNet Subgraph Encoding of Road Segments**

*2.2.2* ***Establishment of Interactive Relationship between Trajectory and Road Network Scene after Scene-Relative Processing.*** Each trajectory point determines its association relationship with each road segment based on its own position and the buffer of each road segment in the road network scene, as shown in Figure 8 below. Let the road network-trajectory association matrix rtadj-matrix be initially initialized as $\{0\}^{Ntr\times Nroad}$, where $Ntr$ is the number of trajectory points and $Nroad$ is the number of road segments in the road network scene. If trajectory point i falls within the buffer of the road segment set$\{r_1、r_2、\cdots、r_n\}$, then set $rtadj_matrix_{i,r_j} = 1/n$; if trajectory point i does not fall within the buffer of any road segment, this point is recorded as an outlier.

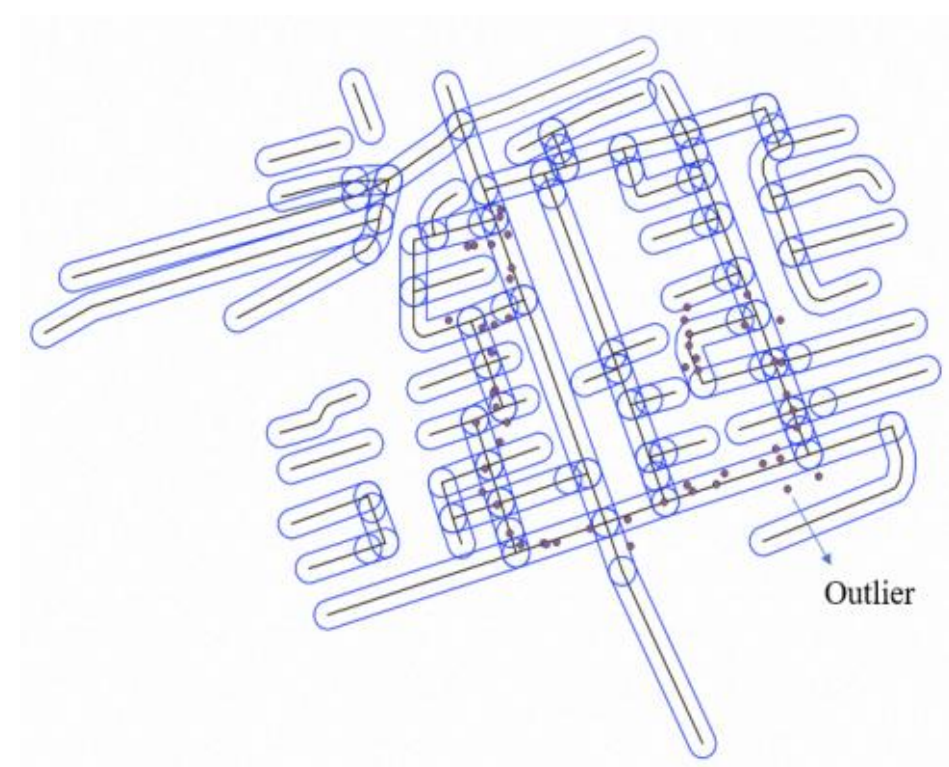

**Figure 8: Association Relationship between Trajectory and Road Segments**

*2.2.3* ***Embedding of Road Network and Trajectory.*** The three types of data after interactive processing are summarized in Table 2 below, and the relationships between these three types of data are shown in Figure 9 below.

**Table 2. Detailed Content of 3 Types of Data**

| | | |
|---|---|---|
| **Road Network Scene Graph Data** | f_nodeList | Node Features of Road Network Scene Dual-Graph |
| | *edge_index* | *Dual-Graph Edge Index (Road Segment Adjacency Relationship)* |
| **Trajectory Graph Data** | tr_edge_index | Trajectory Point Edge Index (Trajectory Point Adjacency Relationship) |
| | *tr_edge_weight* | *Trajectory Point Edge Transition Weight (Default: All 1)* |
| **Dual-Graph Interactive Data** | rtadj_matrix | Trajectory and Road Segment Association Weight Matrix |
| | *sg_mask* | *Outlier Trajectory Mask* |

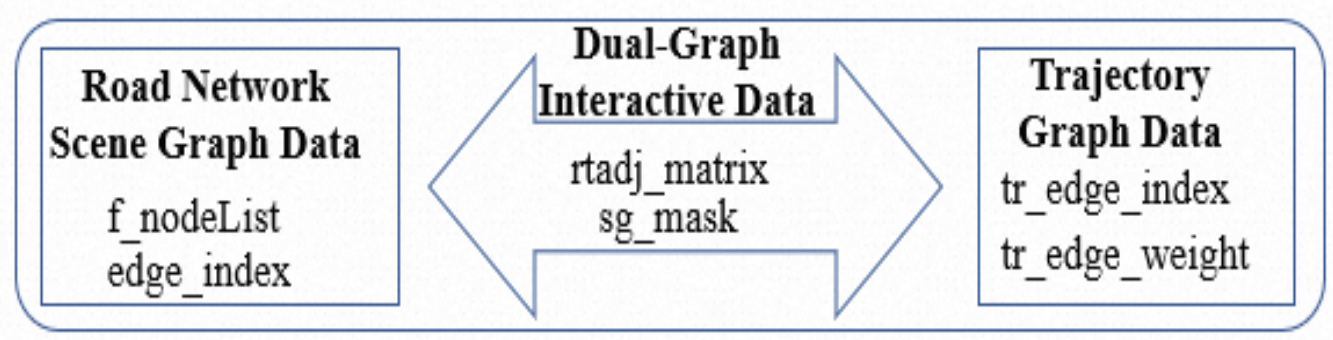

**Figure 9: Interactive Relationship between Types of Data**

Based on the above three types of data, a Graph Neural Network (GNN)-based interactive embedding framework is proposed, as shown in Figure 10, to uniformly model the static topology of the road network and the dynamic behavior of the trajectory. Through the collaborative modeling of the road network dual-graph and the trajectory graph, the embedding of road network scenes and trajectory data is achieved. Through multi-level graph convolution modules, the road network topology and trajectory transition relationships are dynamically fused, finally generating a high-dimensional feature representation that combines structural semantics and dynamic characteristics.

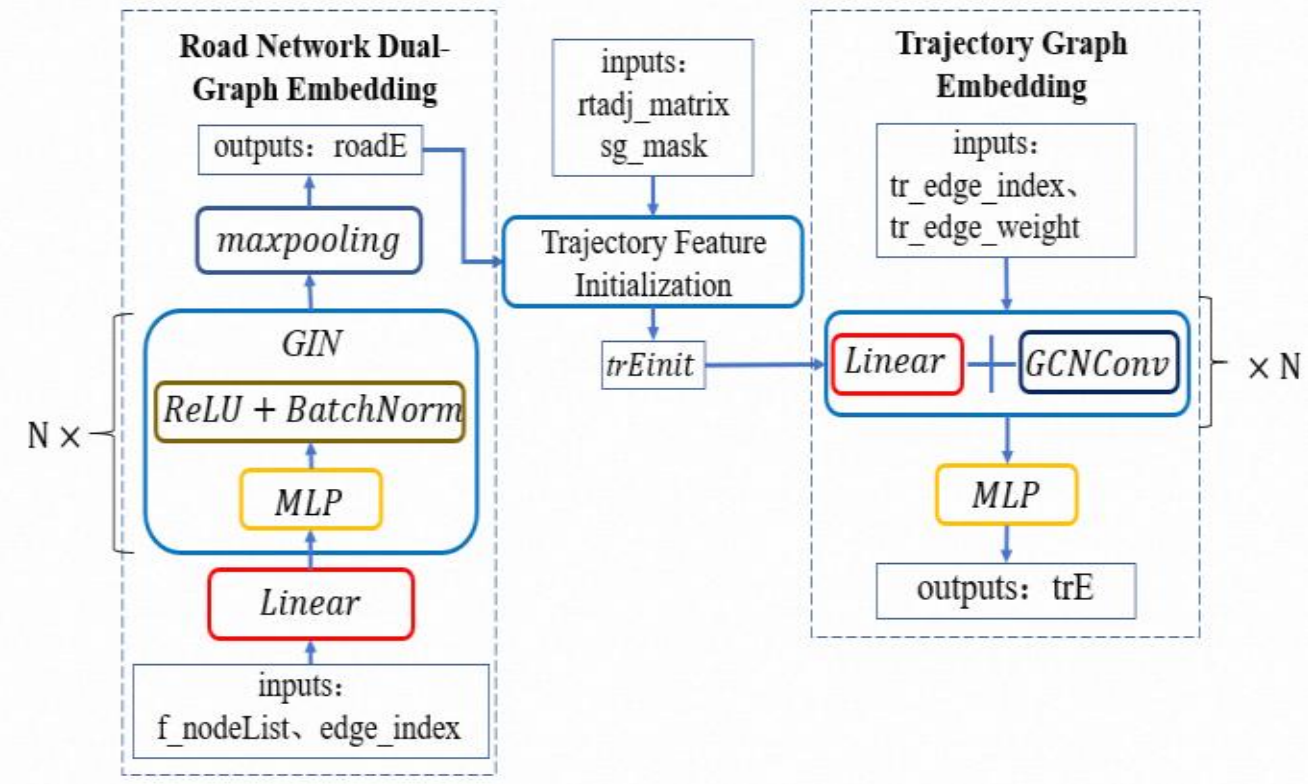

**Figure 10: GNN-based Dual-Graph Interactive Embedding Framework**

**(1) Road Network Dual-Graph Embedding**

A multi-layer Graph Isomorphism Network (RoadGIN) is employed to extract high-order topological features of the road network through the following algorithm:

---

**Algorithm 2:** Road Network Graph Isomorphism Network (RoadGIN)

**Input:** Node features $f_{\text{nodeList}} \in \mathbb{R}^{N_{\text{road}} \times 72}$, edge index edge_index $\in \mathbb{R}^{2 \times N_{\text{edge}}}$, embedding dimension dim, network depth depth.

**Output:** Road network embedding roadE $\in \mathbb{R}^{N_{\text{road}} \times \text{dim}}$.

// Initial node features

1 $h^0 \leftarrow f_{\text{nodeList}}$;

// Step 1: Feature Mapping

// Map original features to embedding space

2 $h^0 \xrightarrow{\text{Linear(72-dim)}} h^0 \in \mathbb{R}^{N_{\text{road}} \times \text{dim}}$;

// Step 2: Multi-layer GIN Convolution

3 **for** $l \leftarrow 1$ **to** *depth* **do**

// Differentiable aggregation via GINConv

4 $h_i^l \leftarrow \text{ReLU}\Big(\text{BatchNorm}\big(\sum_{j \in N(i)} \text{MLP}^l(h_j^{l-1})\big)\Big)$;

5 **end**

// Step 3: Feature Integration

// Fuse multi-layer outputs

6 $\text{roadE}_i \leftarrow \text{maxpooling}(h_i^0, h_i^1, \ldots, h_i^{\text{depth}}, \text{dim} = 0)$;

7 **return** *roadE*;

---

**(2) Trajectory Graph Embedding**

Firstly, the association weight matrix is utilized to achieve soft allocation of road network features to trajectories, supporting many-to-many interactions. An outlier correction mechanism ensures that trajectories not covered by the road network buffer can still obtain effective representations through local feature learning; secondly, spatiotemporal transition patterns of the trajectory are captured via GNN. The overall algorithm is as follows:

---

**Algorithm 3:** Trajectory Feature Learning Algorithm

**Input:** Trajectory and road network association weight matrix $rtadj_matrix \in \mathbb{R}^{N_{tr} \times N_{road}}$, road network embedding $roadE \in \mathbb{R}^{N_{road} \times dim}$, outlier trajectory mask $sg_mask$, edge index $tr_edge_index \in \mathbb{R}^{2 \times edgeCount}$, trajectory point edge transition weight $tr_edge_weight \in \mathbb{R}^{edgeCount}$, embedding dimension $gdim$, and graph convolution depth $depth$.

**Output:** Trajectory features $trE \in \mathbb{R}^{N_{tr} \times gdim}$.

// Step 1: Road Network Feature Propagation

// Map road network embedding to trajectory via matrix multiplication

1 $trEinit \leftarrow rtadj_matrix \times roadE$;

// Step 2: Outlier Correction

// Replace outlier features using MLPstrace on nearest non-outlier points

2 $trEinit_{sg_mask} \leftarrow$ $\text{maxpooling}\left(\sum_{j \in N_A(i)(sg_mask)} \text{MLPstrace}(gridEpre_j), dim = 0\right)$;

// Step 3: Multi-layer GCN Convolution

// Model trajectory dynamic characteristics via linear transformation and GCN

3 $trE^0 \leftarrow trEinit$;

4 **for** $l \leftarrow 1$ **to** *depth* **do**

5 $trE^l \leftarrow \text{GCN}(trE^{l-1}, tr_edge_index, tr_edge_weight)$;

// Step 4: Multi-Layer Perceptron for Trajectory Feature Capture and Summarization

// Final feature extraction using MLPgcn based on gdim

6 $trE \leftarrow \text{MLPgcn}(trE^{depth}, gdim)$;

7 **return** $trE$;

---

## 2.3 Transformer-based Alignment of Road Network Scenes and Corresponding Trajectories

*2.3.1* ***Feasibility and Technical Advantages of Transformer in Map Matching.*** Through its powerful capabilities in sequence modeling, context awareness, and transfer learning, the Transformer provides an end-to-end, data-driven, and context-aware solution for map matching. The application of Transformers in map matching offers significant feasibility and technical advantages, primarily reflected in the following aspects:

### Feasibility:

- Sequence Modeling Capability Adapts to Trajectory Characteristics: The self-attention mechanism of the Transformer excels at handling long-range dependencies in sequence data. Trajectories are essentially spatiotemporal sequence data, whose motion patterns (such as path selection and turning decisions) exhibit complex spatiotemporal correlations. The process of mapping trajectory point sequences to the road network can be analogous to the "Machine Translation" task in Natural Language Processing: translating trajectory point sequences (source language) into road segment sequences (target language). This analogy allows the sequence modeling capability of the Transformer to be directly transferred to the map matching scenario;

- Scalability of Context-aware Modeling: The Transformer can automatically learn driver behavior patterns (such as preferring specific road grades and avoiding congested segments) and road network structure features (such as intersection connectivity) through large-scale trajectory data. This data-driven approach is particularly suitable for capturing complex features such as "driver shared preferences" and "spatial noise patterns".

**Technical Advantages:**

- Modeling of Complex Spatial Dependencies: Through the multi-head attention mechanism, the Transformer can simultaneously capture global spatial dependencies between trajectory points (such as path selection across multiple intersections) and local temporal dynamics (such as short-term turning behavior). This capability makes it superior to traditional methods based on local proximity matching (such as nearest road point search) or local optimal paths (such as dynamic programming);
- Context-aware Noise Robustness: Traditional methods are sensitive to trajectory noise. Through self-attention at the encoder end, the Transformer can automatically learn noise patterns (such as positioning drift patterns in specific areas) and perform joint optimization combined with road network topology at the decoder end;
- Scalability for Large-scale Trajectory Data: The parallel computing characteristics of the Transformer make it more suitable for processing massive trajectory data. Deep learning-based frameworks can continuously improve performance after training on large-scale data, facilitating expansion to super-large urban road networks.

*2.3.2* ***Transformer-based Road Network Scene Spatial Alignment Framework.*** Combining the dual-graph interactive embedding results of trajectories and road network scenes, this paper proposes a Transformer-based Road Network Scene Spatial Alignment Framework. Through customized feature encoding, road network embedding decoder input, and CRF(Conditional Random Fields) joint optimization, the framework utilizes the attention mechanism of the Transformer to achieve fine-grained alignment between trajectory points and road network features, extending the Transformer to the trajectory-road network spatial alignment task while combining global modeling capabilities with domain knowledge fusion. This framework employs road network embeddings as a dynamic candidate library and utilizes dot product similarity to replace the traditional Softmax, effectively solving the road network scene spatial alignment problem. The CRF layer explicitly models topological constraints between roads (such as connectivity and turning restrictions), compensating for the Transformer's deficiency in local transition patterns.

*2.3.3* ***Core Computational Steps.*** Similar to the computational process of traditional Transformers, the core computational steps are as follows:

**Algorithm 4:** Transformer-based Traj-to-Road Mapping Algorithm

**Input:** Trajectory sequence $T = \{t_1, t_2, \ldots, t_L\}$,
Road network embeddings $roadE \in \mathbb{R}^{N_{road} \times d_{model}}$,
Learnable start token embedding $start_emb$,
Transition matrix for road connectivity constraints.

**Output:** Predicted road sequence probabilities and optimized predictions.

```
   // Step 1: Input Projection and Positional Encoding
   // Enhance trajectory embeddings with temporal information
1  H^0 ← Linear(T) + PositionalEncoding(L);
   // Step 2: Encoder Context Generation
   // Encode trajectory sequence into context representation with global dependencies
2  Memory ← TransformerEncoder(H^0);
   // Step 3: Decoder Generation of Step-wise Predicted Road Probabilities
   // Autoregressive decoding with causal mask
3  Y_0 ← start_emb;
4  for i ← 1 to L do
       // Generate causal mask to prevent future information leakage
5      Mask_i ← CausalMask(i);
       // Cross-attention between decoder and encoder memory
6      H_i^dec ← TransformerDecoder(Y_{i-1}, Memory, Mask_i);
       // Road candidate matching via dot product similarity
7      prob_i ← Softmax(H_i^dec · roadE^T);
8      Y_i ← Y_{i-1} ⊕ argmax(prob_i);
   // Step 4: CRF Structured Prediction
   // Optimize predictions using road connectivity constraints
9  probs ← [prob_1, prob_2, ..., prob_L];
10 final_predictions ← CRF(probs, TransitionMatrix);
11 return final_predictions;
```

The differences between the proposed framework and the traditional Transformer are primarily the following four points, as shown in Table 3 below.

**Table3. Differences Between Our Framework and the Traditional Transformer**

| Dimension | Traditional Transformer | Our Framework |
|---|---|---|
| Output Space Design | Directly predicts classes in a fixed vocabulary (e.g., tokens) | Computes similarity between decoder outputs and road network embeddings via dot-product for dynamic road candidate matching |
| Decoder Input Construction | Fixed word embedding table + positional encoding | Road network embeddings act as a dynamic candidate repository; step-wise input is the road embedding predicted in the previous step |
| Task-Specific Masking | Only padding masks | Introduces invalid road segment masks to precisely control attention scope |
| Structured Prediction | Independent Softmax classification | Integrates a CRF layer to model road transition constraints, enhancing sequence connectivity validity |

# 3 Experimental Evaluation

## 3.1 Training Performance Test in Conformal Unit Disk Coordinates

To verify the impact of the conformal mapping-based scene relativization strategy on model training efficiency and stability, this paper compares the training processes under two input coordinate systems with identical network architectures and hyperparameter settings: (1) the original geographic coordinate system (WGS-84 latitude/longitude or projected plane coordinates); and (2) the unit disk local relative coordinate system transformed via conformal mapping. The experiments are conducted on Jinan Dataset A (positioning drift variance: 16 m; sampling step size: 30 m; 8,352 samples), utilizing the full SceneGTMM framework for end-to-end training. The training loss and validation accuracy for each of the first 20 epochs were recorded.

As shown in Figure 11, under the unit disk coordinate system, the model training loss decreases more rapidly with significantly reduced fluctuations. Specifically, training with original geographic coordinates exhibits severe loss oscillations within 20 epochs, whereas the loss curve under the unit disk coordinate system is smooth and converges to a stable plateau within approximately 20 epochs. In contrast, the original coordinate system fails to achieve comparable performance within the same 20 epochs. Furthermore, validation accuracy (AC) improves more rapidly under the unit disk coordinate system, reaching 0.95 as early as the 13th epoch, whereas the original coordinate system achieves only 0.85 at the same stage.

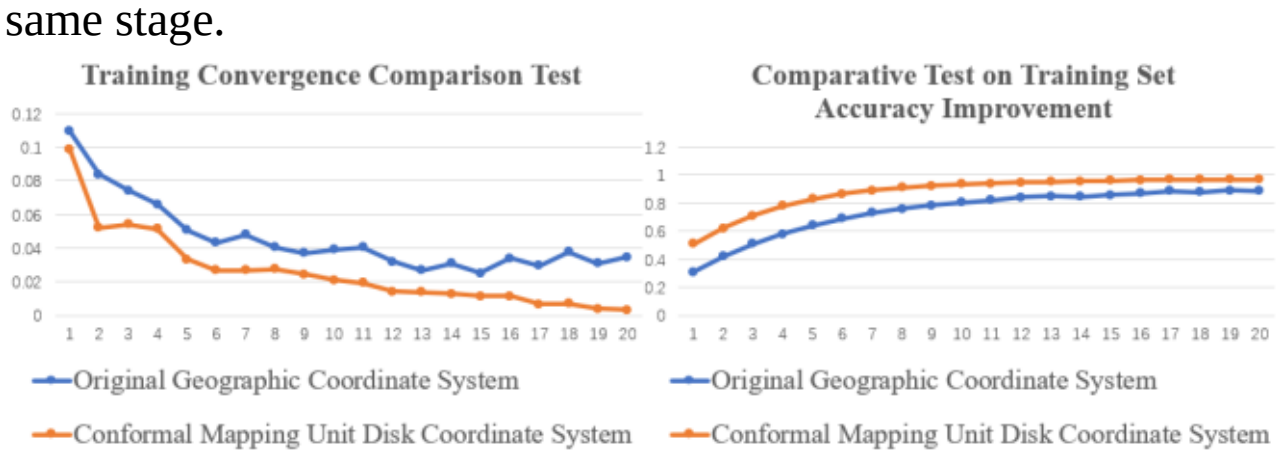

**Figure 11: Training Performance Comparison: Original Coordinates vs. Conformal Mapping Unit Disk Coordinates**

This phenomenon can be explained from the following three perspectives:

**(1) Scale Normalization and Enhanced Numerical Stability**

Original geographic coordinates are influenced by absolute positions (e.g., significant differences in coordinate values between city centers and suburbs), resulting in severe distribution shifts in input features across different trajectory samples. Conformal mapping compresses each trajectory's local road network into a unit disk, representing all samples within a unified scale (radius ≤ 1) and a centrally symmetric space. This significantly alleviates the heterogeneity of input distributions and improves the stability of gradient propagation, thereby avoiding numerical overflow or gradient explosion caused by large-scale coordinate values.

**(2) Conformality of Geometric Structure Facilitates Topology Perception**

The core property of conformal mapping is "conformality," meaning that key geometric relationships such as local road intersection angles and turning directions are preserved after mapping. This property enables the GNN to more accurately capture the semantic consistency of road connections (e.g., T-junctions, roundabouts) when aggregating neighborhood information, avoiding topological misjudgments caused by coordinate distortion. Meanwhile, the Transformer also benefits from the preservation of angular structures when calculating relative positions between trajectory points, thereby learning turning behavior patterns more efficiently.

**(3) Elimination of Absolute Position Dependency and Focus on Scene Semantics**

In the original coordinate system, the model tends to over-rely on absolute positions (e.g., "a certain latitude/longitude must correspond to a specific main road"), leading to limited generalization capability. In contrast, the unit disk coordinate system constructs a local reference frame centered on the current trajectory, forcing the model to make decisions based solely on relative geometric and topological structures, essentially achieving "ID-removal" and "scene-awareness." This representation not only reduces model capacity requirements (no need to memorize global road network IDs) but also focuses the optimization objective more on learning general matching rules, thereby accelerating convergence and improving generalization performance.

In summary, conformal mapping to the unit disk is not merely a coordinate transformation but also an effective means of feature normalization and structural regularization. Through the three major mechanisms of geometric conformality, scale normalization, and scene centering, it significantly improves the training efficiency and convergence stability of SceneGTMM, laying a solid foundation for subsequent cross-regional transfer and dynamic road network adaptation.

## 3.2 Applicability Test of the Model on Traj-ectory Data with Varying Localization Accu-racy

A region in Jinan City, Shandong Province, was selected as the experimental dataset. The region contains three trajectory datasets with distinct localization accuracies, As shown in Table 4.

**Table 4. Three Trajectory Datasets**

| Dataset | Drift variance | Sampling Interval |
|---|---|---|
| A | 16.0 meters | 30.0 meters |
| B | 26.0 meters | 60.0 meters |
| C | 50.0 meters | 30.0 meters |

The three datasets were partitioned into training and test sets at an 8:2 ratio to train and evaluate the model's adaptability under varying positioning accuracies. Figure 13 illustrates the model's inference results across the three datasets. The results demonstrate that model training and inductive inference, incorporating scene relativization and dual-graph interaction map matching, were successfully conducted on trajectory data with varying positioning accuracies. For each of the three datasets, 100 trajectories from the test set were selected for inference testing; the corresponding matching accuracies are listed in Table 5. Although the matching accuracy decreases slightly as the positioning accuracy of the datasets declines, it remains high overall. This indicates that the model's dual-graph interaction mode can effectively learn noise patterns. In other words, stable positioning noise has a relatively minor impact on the model's matching accuracy.

**Table 5. Matching Accuracy Comparison by Dataset**

| Dataset | AC | JS | BLEU |
|---|---|---|---|
| A**(Tra + Inf)** | 0.8416 | 0.9259 | 0.8233 |
| B**(Tra + Inf)** | 0.8289 | 0.9167 | 0.8174 |
| C**(Tra + Inf)** | 0.8325 | 0.9189 | 0.8192 |

## 3.3 Robustness Testing for Model Training and Inductive Inference

Datasets A, B, and C from the previous section served as the experimental data. Tests were conducted across the following categories, with the matching accuracies for each category presented in Table 6.

**Table 6. Matching Accuracy Comparison across Seven Test Categories**

| Test Category | Tra Dataset | Inf Dataset | AC | JS | BLEU |
|---|---|---|---|---|---|
| 1 | A、B、C | A | 0.8277 | 0.8862 | 0.8124 |
| 2 | A、B、C | B | 0.8266 | 0.9015 | 0.8011 |
| 3 | A、B、C | C | 0.8352 | 0.8923 | 0.8213 |
| 4 | A | A | 0.8416 | 0.9259 | 0.8233 |
| 5 | A | C | 0.7128 | 0.7978 | 0.7156 |
| 6 | C | A | 0.8113 | 0.8566 | 0.8055 |
| 7 | C | C | 0.8325 | 0.9189 | 0.8192 |

(1) For Categories 1-3 (mixed training, with inference on A, B, and C respectively), the matching accuracy remains at a high level. This suggests that mixed training enables the model to learn shared features across different datasets (e.g., road network structure and the spatial relationship between trajectories and the road network), thereby maintaining stable performance on trajectory data with varying drift variances and sampling step sizes.

(2) Categories 5 (trained on A → inferred on C) and 6 (trained on C → inferred on A) demonstrate that the model exhibits better transfer adaptability from "high noise to low noise" than from "low noise to high noise," which aligns with intuitive expectations. The accuracy of Category 6 is comparable to that of Category 1 (mixed training), indicating that a model trained on high-drift data (C) can adapt well to low-drift data (A). The high-noise characteristics of C compel the model to learn more robust road network matching rules. Although the accuracy of Category 5 decreases, it drops by only about 15% even when the drift degree of the test set exceeds 50%, indicating strong generalization capability.

(3) The accuracy of Category 7 (trained on C → inferred on C) is slightly lower than that of Category 3 (mixed training), indicating that mixed training does not compromise the optimal performance achievable on a single dataset while simultaneously enhancing cross-dataset stability.

In summary, the Conformal Mapping Scene Relativization Strategy and the GNN-Transformer Dual-Graph Interaction Framework effectively integrate the topological relationships between trajectories and the road network. They demonstrate strong robustness across trajectory data with varying positioning accuracies and sampling densities, exhibiting particularly stable performance in high-drift or sparse trajectory scenarios.

## 3.4 Comparative Experiment on Model Transferability

**Baseline Models**

- HMM [2]: This method recovers position information based on the assumption of linear and uniform trajectory motion and conducts trajectory matching via the road network. The Hidden Markov Model (HMM) has demonstrated superior accuracy in high-sampling-rate trajectory map matching;
- MTrajRec [4]: Utilizing a multi-task sequence-to-sequence learning architecture, this model simultaneously predicts road segments and movement ratios. It incorporates constraint masks, attention mechanisms, and attribute modules to overcome the limitations of coarse-grained grid representation and enhance performance;
- GraphMM [7]: This model leverages the graph characteristics of map matching, combining Graph Neural Networks with conditional models. It fully exploits the graph topological structure of both the road network and trajectories, aligning road segments and trajectories within the latent space;
- TMM-LGD [9]: This method employs a Transformer-based encoder-decoder map matching framework. It automatically learns contextual representations of noisy trajectory points in an end-to-end manner, inferring driver trajectory behavior and road network structure.

This study selects HMM, MTrajRec, GraphMM, TMM-LGD, and SceneGTMM to compare matching accuracy in regions within the training dataset and regions outside the training dataset. The results are presented in Table 7.

**Table 7. Matching Accuracy Evaluation of All Algorithms**

| Algorithms | Tra-Dataset Region Inf | | | Transfer to Non-Tra-Dataset Region Inf | | |
|---|---|---|---|---|---|---|
| | AC | JS | BLEU | AC | JS | BLEU |
| HMM | 0.7912 | 0.8599 | 0.8033 | 0.8041 | 0.8683 | 0.8098 |
| MTrajRec | 0.8281 | 0.8755 | 0.8197 | **Cannot achieve transfer reasoning** | | |
| GraphMM | 0.8414 | 0.9123 | 0.8219 | **Cannot achieve transfer reasoning** | | |
| TMM-LGD | 0.8461 | 0.9285 | 0.8399 | **Cannot achieve transfer reasoning** | | |

| SceneGTMM | **0.8554** | **0.9371** | **0.8487** | **0.8222** | **0.8917** | **0.8332** |
|---|---|---|---|---|---|---|

(1) During inference within training dataset regions, MTrajRec, GraphMM, TMM-LGD, and SceneGTMM all exhibit high matching accuracy (attributable to the small size of the experimental region and the high sampling frequency); however, SceneGTMM achieves higher accuracy than the other algorithms. The higher matching accuracy of TMM-LGD and SceneGTMM compared to MTrajRec and GraphMM reflects the advantage of the Transformer framework in preserving information over long distances during matching. SceneGTMM inherits the Transformer's capability for contextual representation of noisy trajectory points, outperforming traditional HMM and the sequence-to-sequence architecture-based MTrajRec. Furthermore, its deep modeling of road network topology via a Transformer + Graph Neural Network combination is more comprehensive than that of GraphMM (which uses only GNN) and TMM-LGD (which uses only Transformer).

(2)In transfer inference to non-training dataset regions, MTrajRec and TMM-LGD fail to achieve transfer inference because their model training is bound to the road network of the training set. Similarly, GraphMM fails because its trajectory-enhanced graph is bound to the training set road network. In inference within non-training dataset regions, SceneGTMM is the only deep learning model capable of transfer, with performance significantly superior to traditional HMM. The Scene Relativization Strategy achieves cross-region transfer by dynamically adjusting the relative feature representation of trajectories and the road network, thereby reducing dependency on the training region's road network.

In summary, SceneGTMM, leveraging the GNN-Transformer Dual-Graph Interaction Framework and the Conformal Mapping Scene Relativization Strategy, resolves the dependency of existing methods on the training region's road network while maintaining the high accuracy of deep learning models, thereby achieving cross-region transfer capability. It achieves optimal performance in training regions and is significantly superior to the traditional HMM method in non-training regions, embodying a combination of theoretical innovation and engineering practicality.

## 4 Conclusion

Serving as a bridge between the physical space and the digital world, the accuracy and real-time performance of map matching directly impact the development of Intelligent Transportation Systems (ITS), autonomous driving, and smart cities. The SceneGTMM framework proposed in this paper leverages the Scene Relativization Strategy and the GNN-Transformer Dual-Graph Interaction Architecture to overcome bottlenecks in transferability, dynamic adaptability, and interpretability faced by traditional methods and deep learning models. It offers a novel solution for map matching in scenarios characterized by high noise, cross-region operations, and dynamic road networks.

First, the Scene Relativization Strategy effectively decouples the model from the training road network by dynamically constructing local coordinate systems and employing road network pruning mechanisms, significantly enhancing cross-region transfer capability. Experiments demonstrate that during inference in non-training regions, SceneGTMM achieves a matching accuracy (AC=0.8222) that surpasses the traditional HMM model (AC=0.8041), verifying its dynamic scene adaptation capability. Second, the GNN-Transformer Dual-Graph Interaction Architecture achieves a deep fusion of local topological constraints and global path preferences. The GNN models the local geometric features and connectivity of the road network, while the Transformer captures the long-range dependencies of trajectory points. Furthermore, the cross-graph attention mechanism enhances the model's robustness to multi-path effects through visualized association weights. The introduction of the CRF layer further optimizes the topological coherence of road sequences, addressing the limitation of traditional Transformers in modeling local transition rules.

Experimental results indicate that SceneGTMM maintains stable performance (AC=0.8124-0.8352) across trajectory data with varying positioning accuracies (16-50m drift variance) and sampling densities (30-60m step size). Moreover, in transfer scenarios from high noise to low noise (e.g., trained on C → inferred on A), the accuracy decreases by only 2.4%, highlighting its generalization capability in learning noise patterns. Compared to existing deep learning models (such as MTrajRec, GraphMM, and TMM-LGD), SceneGTMM is the only framework that supports dynamic road network updates and cross-region transfer, offering the potential for practical engineering deployment in real-time map matching.

Future research directions include: 1) Introducing multi-modal data (such as traffic flow and traffic signal status) to enhance dynamic path planning capabilities; 2) Optimizing the model's lightweight design to adapt to vehicle-mounted edge computing scenarios; 3) Exploring interpretability enhancement mechanisms based on neuro-symbolic systems, combining the advantages of rule-based knowledge and deep learning. The proposal of SceneGTMM not only advances map matching technology but also provides a new paradigm for the deep integration of spatio-temporal perception and Intelligent Transportation Systems.

## Acknowledgements

We thank all members of the Geometry One Map project team at Amap, Alibaba Group for their insightful discussions and cooperation in data acquisition.

## A Transformer-based Road Network Scene Spatial Alignment Framework

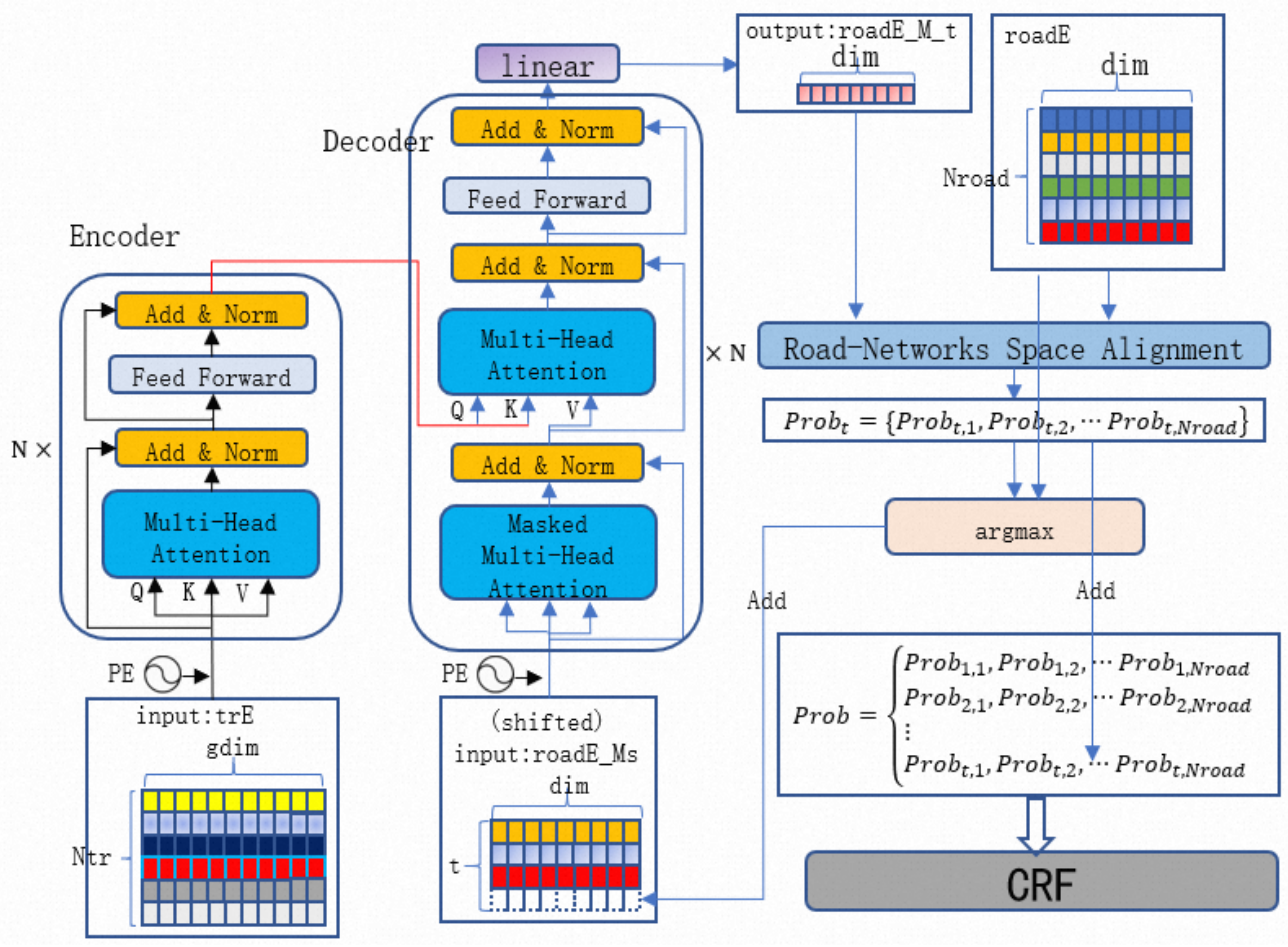

**Figure 12.Transformer-Based Spatial Alignment Framework for Road Network Scenes**

## B CRF[21、22] Loss Function

The integration of CRF and Transformer in trajectory matching tasks enables dual constraints: Transformer captures long-range dependencies (e.g., trajectory detours) while CRF supplements local transition rules. The synergistic interaction between CRF and Tran-sformer is as follows: Transformer extracts global depe-ndencies between trajectories and road networks, outputting emission scores $f(y_t, x_t)$ at each step, with transition matrices dynamically generated from scenario-adapted road network data for each sample; CRF then optimizes structured road sequence predictions based on these emission scores and transition matrices to ensure physical connectivity.

In trajectory matching tasks, CRF acts as a structured decision-making engine, addressing label dependency issues by enforcing physical connectivity of road sequences through transi-tion matrices, compensating for Transformer's limitations in capturing local transition patterns to correct misaligned segments, and enabling end-to-end optimization. This joint training with Transformer achieves globally optimal tr-ajectory-road network alignment, where the CRF loss function backpropagates gradients to optimize both Transformer parameters and dual-graph interactive embeddings.

### 1. Fundamentals of CRF

CRF is a discriminative probabilistic grap-hical model specifically designed for sequence labeling tasks. Its core principle is to model transition constraints between labels through a transition matrix. Unlike independent classify-ers (e.g., Softmax), CRF explicitly learns depe-ndencies between labels by incorp-orating these transitions, such as the connectivity of road sequences in map matching.

Mathematical Definition: Given an input sequence $X = (x_1, x_2, \cdots, x_T)$ and a label seq-uence $Y = (y_1, y_2, \cdots, y_T)$, the conditional probability of a Conditional Random Field (CRF) is defined as:

$$P(Y|X) = \frac{1}{Z(x)} \exp\left(\sum_{t=1}^{T} (f(y_t, x_t) + g(y_{t-1}, y_t))\right) \quad (2)$$

Where: $f(y_t, x_t)$: Emission score, quantify-ing the contribution of input $x_t$ to label $y_t$ (supplied by the Transformer decoder output in this work); $g(y_{t-1}, y_t)$: Transition score, representing the transiti-on probability from labe $y_{t-1}$ to $y_t$; $Z(x)$: Normalization factor, ensuring the total probability over all label sequences sums to 1.

### 2. CRF Implementation Details in Trajectory-Road Network Alignment

(1) Loss Function Computation

Negative Log-Likelihood Loss: For a grou-nd-truth label sequence $Y^*$, the loss function is defined as:

$$L = -\log\left(\exp\left(S(Y^*)\right)/Z(x)\right) \quad (3)$$

Where: $S(Y) = \sum_{t=1}^{T} (f(y_t, x_t) + g(y_{t-1}, y_t))$: Sequence score; $Z(x) = \sum_{Y'} \exp\left(S(Y')\right)$: Norma-lization factor computed via dynamic program-ing (forward algorithm), avoiding exhaustive enumeration of all possible label sequences.

(2) Decoding the Optimal Label Sequence

Viterbi Algorithm: Based on emission scor-es and the transition matrix, dynamic program-mming searches for the globally optimal label sequence:

$$Optimal\ Path\ Score = \max_{Y} \sum_{t} (f(y_t, x_t) + g(y_{t-1}, y_t)) \quad (4)$$

(3) Emission Matrix and Transition Matrix

The Emission Matrix $f(y_t, x_t)$, which captures global dependencies between trajecto-ries and the road network through Transformer-based feature extraction;

The Transition Matrix $g(y_{t-1}, y_t)$, enco-ding the topological relationships (e.g., connec-tivity, directional constraints) between road segments in the scenario-specific road network.

## C Accuracy Evaluation Methods

Model performance is evaluated using three metrics: Accuracy (AC), Jaccard Similarity (JS), and Bilingual Evaluation Understudy (BLEU).

Let $T_i = \{R_{i1}, R_{i2}, \cdots, R_{iL}\}$ and $\{\hat{R}_{i1}, \hat{R}_{i2}, \cdots, \hat{R}_{iL}\}$ denote the ground-truth and predicted paths for the i-th trajectory $P_i$, where $R_{ij}$ and $\hat{R}_{ij}$ represent road segments, and $iL = |T_i| = |\hat{T}_i|$ denotes the number of segments in each path. The metrics are defined as follows:

(1) Accuracy(AC)

The mathematical definition of AC is provided in Equations (5) and (6). This metric calculates the ratio of intersecting road segments between the predicted and ground truth paths to the total number of road segments in the ground truth path, averaged over all trajectories.

$$AC = \frac{1}{N} \sum_{i=1}^{N} \frac{\sum_{j}^{iL} \pi(R_{ij} = \hat{R}_{ij})}{|T_i|} \quad (5)$$

$$\pi(A) = \begin{cases} 1, A成立 \\ 0, A不成立 \end{cases} \quad (6)$$

(2) Jaccard Similarity (JS)

The mathematical definition of JS is provided in Equation (7). It measures the similarity between the predicted and ground truth paths by calculating the ratio of the number of intersecting road segments to the number of union road segments, with a value range from 0 (no common segments) to 1 (perfect match).

$$JS = \frac{1}{N}\sum_{i=1}^{N}\frac{|T_i \cap \hat{T}_i|}{|T_i \cup \hat{T}_i|} \quad (7)$$

(3) Bilingual Evaluation Understudy (BLEU)

BLEU is a metric commonly employed to evaluate the quality of machine-generated text. In this paper, BLEU is utilized to assess the similarity between predicted matching paths and ground truth paths. Its core idea is to measure the degree of matching by comparing n-grams (sequences of nn consecutive road segments) within the predicted and ground truth paths (specifically, the equally weighted average of n-grams with nn ranging from 1 to 4 in this paper). The core logic is that superior path matching should exhibit n-gram statistical features similar to those of the ground truth paths. BLEU (Equation 10) consists of two components: Precision (Equation 8) and Brevity Penalty (Equation 9). Equation 8 counts the number of matching consecutive n-grams between the candidate path and the reference path and divides it by the total number of n-grams in the candidate path; Equation 9 imposes a penalty on cases where the predicted path is too short (reducing the score via an exponential function when the predicted path length is less than the ground truth path length). This design focuses on both the matching degree of local patterns (Precision) and prevents the model from "cheating" by overly shortening the predicted path (Brevity Penalty).

$$\Pr(T_i, \hat{T}_i) = \frac{\text{Number of matching } n\text{−grams}}{\text{Total number of } n\text{−grams}} \quad (8)$$

$$BP_i = \begin{cases} 1 & |\hat{T}_i| > |T_i| \\ \exp\left(1 - \frac{|\hat{T}_i|}{|T_i|}\right) & |\hat{T}_i| \le |T_i| \end{cases} \quad (9)$$

$$\text{BLEU} = \frac{1}{N}\sum_{i=1}^{N} BP_i \cdot exp\left(\frac{1}{n}\sum_{j=1}^{n}\log\left(\Pr(T_i, \hat{T}_i)_j\right)\right) \quad (10)$$

## D Matching Performance vs. Sampling Frequency and Positioning Accuracy.

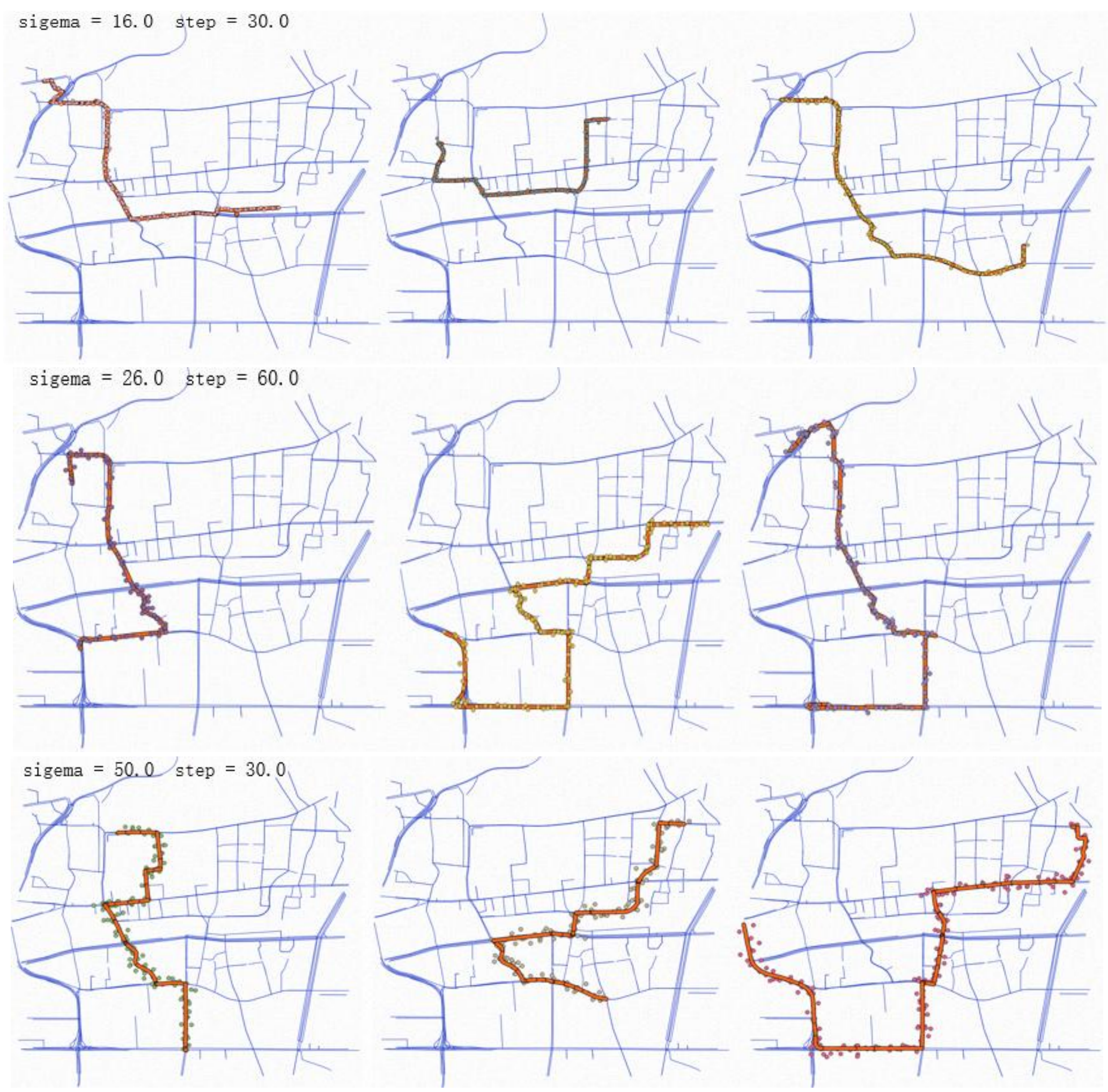

**Figure 13. Matching Performance vs. Sampling Frequency and Positioning Accuracy**